\documentclass[letterpaper, 10 pt, conference]{ieeeconf}  

\IEEEoverridecommandlockouts                              

\usepackage{graphicx}
\usepackage{amsmath}
\usepackage{amssymb}
\usepackage{algorithm}
\usepackage{algpseudocode}
\usepackage[labelformat=simple]{subcaption}
\usepackage[table,xcdraw,dvipsnames]{xcolor}
\usepackage[bookmarks=true,colorlinks=true]{hyperref}
\usepackage{svg}
\usepackage[export]{adjustbox}
\usepackage[normalem]{ulem}
\useunder{\uline}{\ul}{}

\newcommand{\method}{WayFASTER}

\title{\LARGE \bf
\method: a Self-Supervised Traversability Prediction for Increased Navigation Awareness}

\author{Mateus V. Gasparino$^{1}$, Arun N. Sivakumar$^{1}$, and Girish Chowdhary$^{1}$
\thanks{Project repository: \url{https://github.com/matval/wayfaster}}%
\thanks{$^{1}$Field Robotics Engineering and Science Hub (FRESH), Illinois Autonomous Farm, University of Illinois at Urbana-Champaign (UIUC), IL}%
\thanks{{Correspondence to \tt\small \{mvalve2,girishc\}@illinois.edu}}%
\thanks{{This work was supported in part by NIFA \#2021-67021-33449, NIFA \#2021-67021-34418, and AFRI grant \#2020-67021-32799/project accession no.1024178 NSF/USDA National AI institute: AIFARMS.}}%
}

\begin{document}

\maketitle
\thispagestyle{empty}
\pagestyle{empty}

\begin{abstract}

Accurate and robust navigation in unstructured environments requires fusing data from multiple sensors. Such fusion ensures that the robot is better aware of its surroundings, including areas of the environment that are not immediately visible but were visible at a different time. To solve this problem, we propose a method for traversability prediction in challenging outdoor environments using a sequence of RGB and depth images fused with pose estimations. Our method, termed \method{} (Waypoints-Free Autonomous System for Traversability with Enhanced Robustness), uses experience data recorded from a receding horizon estimator to train a self-supervised neural network for traversability prediction, eliminating the need for heuristics. Our experiments demonstrate that our method excels at avoiding obstacles, and correctly detects that traversable terrains, such as tall grass, can be navigable. By using a sequence of images, \method{} significantly enhances the robot's awareness of its surroundings, enabling it to predict the traversability of terrains that are not immediately visible. This enhanced awareness contributes to better navigation performance in environments where such predictive capabilities are essential. 

\end{abstract}

\section{Introduction}

Robots can play a crucial role in performing demanding and hazardous tasks in outdoor environments, including agriculture, construction, mining, and remote area exploration. However, ensuring safe navigation in such challenging terrains requires that the robot can discern traversable paths. 
Heuristic-based methods of determining traversability present considerable difficulties in their design and are prone to generating false positives, particularly in outdoor settings where the appearance of geometric traversability does not always align with practical traversability, such as for tall grass, muddy terrains, and snow \cite{gasparino2022wayfast,kahn2021badgr}.

\begin{figure}[t]
    \centering
    \frame{\includegraphics[trim={6cm 0cm 6cm 0cm},clip,width=0.84\linewidth]{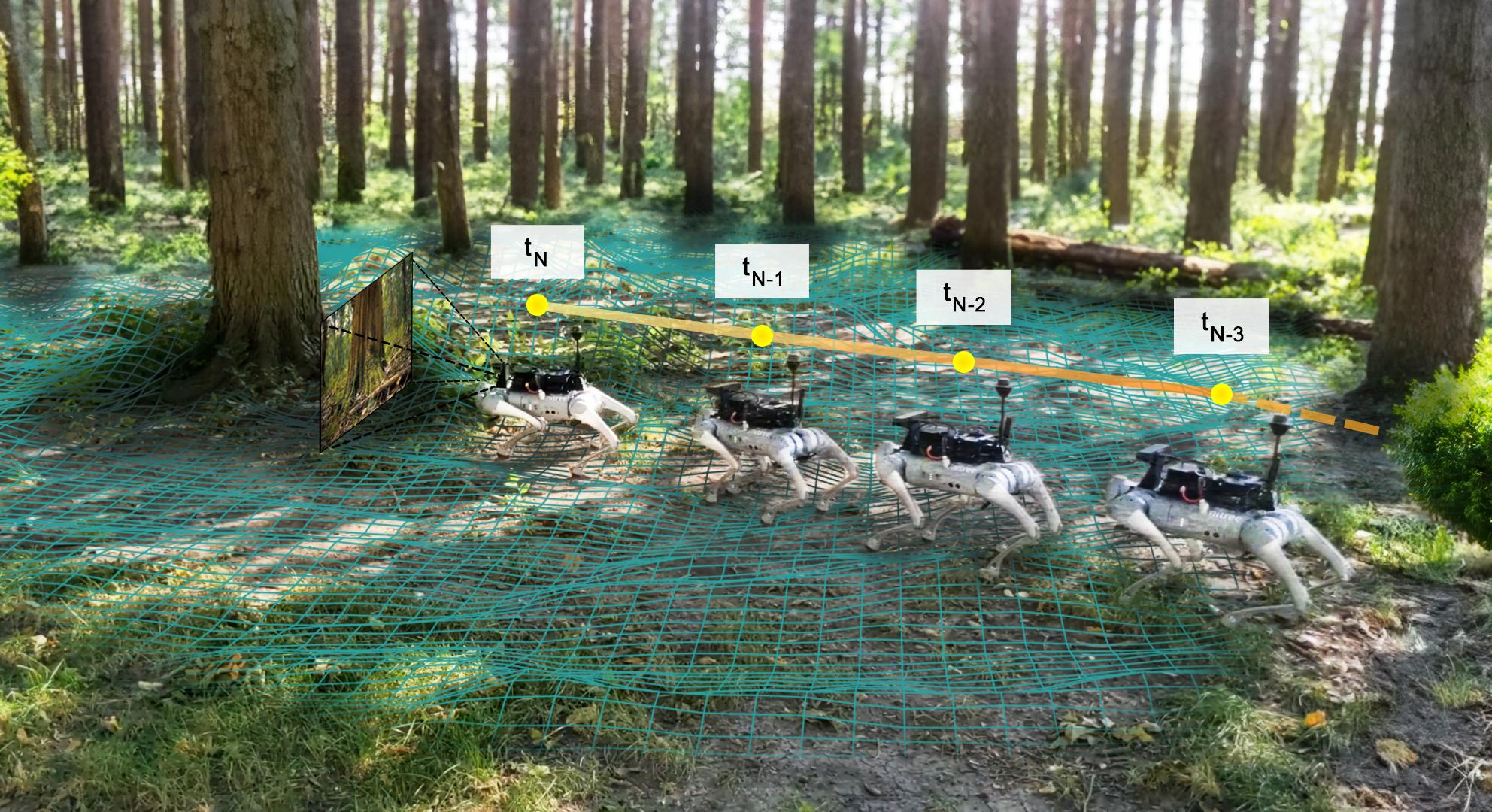}}
    \caption{\textbf{\method} accumulates a sequence of RGB-D images to predict a wide-view traversability map.}
    \label{fig:main-image}
    \vspace{-0.2in}
\end{figure}

Learning-based methods provide a compelling alternative to determine traversability based on the robot's own experience \cite{kahn2021badgr,howard2006towards,kim2006traversability,wellhausen2020safe,miki2022learning}. However, effectively learning traversability can be a big challenge in the supervised-learning setting, since a large amount of annotated data is usually required to train the network models \cite{valada2017adapnet,loquercio2018dronet,chiang2019learning}. Existing methods have tried to avoid the need for manually annotated data by using heuristics such as collisions, robot attitude, or bumpiness to determine traversability labels \cite{kahn2021badgr}. Yet, this essentially heuristic approach is limiting since it relies on a human expert's judgment to define thresholds to what is appropriate for navigation. Moreover, these heuristics are difficult to design and could lead to wrong labels. For example, many untraversable terrains, such as snow, sand, or mud are not bumpy, and bumpiness cost would be easily misinterpreted as traversability. 

Another challenge in the domain of traversability prediction lies in the intricacies of information fusion. Robots have many sensors, and fusing this multi-modal information to create a reliable map is crucial for ensuring secure navigation in complex, unstructured environments. For instance, a single instantaneous camera observation can only capture the information within its immediate field of view \cite{gasparino2022wayfast,kahn2021badgr,wellhausen2020safe}. However, by fusing multiple measurement steps, the resultant map prediction can encompass a far broader spectrum of information, thereby improving the robustness in navigation \cite{hu2021fiery}. This is crucial in instances where an untraversable terrain, such as a ditch, suddenly leaves the robot's field of view. If the robot drives close to this area, deeming it traversable, it could land in the ditch! Hence, an approach that fuses multiple observations to create a broader picture of the robot's surroundings is needed.


\subsection{Background Work}

The landscape of the related work can be broadly clustered into classical methods, (supervised) learning-based methods, and self-supervised methods. 

In \textbf{classical navigation}, waypoints-based path planning is the standard method for robot path programming, which works well in structured outdoor environments where obstacles are known and static \cite{gasparino2023cropnav}. For unstructured environments, classical approaches often do reactive obstacle avoidance through heuristics based on sensor data \cite{howard2001vision,chilian2009stereo,wermelinger2016navigation}. The estimation of traversability has primarily relied on rule-based attributes extracted from geometric and visual characteristics, such as the elevation, roughness, inclination, or predefined visual features of the terrain \cite{wermelinger2016navigation}. In all these approaches, traversability ignores important information that visual semantic information can bring, limiting their performance in challenging unstructured outdoor environments.

In \textbf{Learning-based navigation}, semantic segmentation for categorization of off-road terrains and obstacles is the de facto method \cite{valada2017adapnet,maturana2018real,shaban2022semantic}. Specific to autonomous vehicles, a line of research has emerged to predict semantic maps in the bird's-eye-view (BEV) \cite{hu2021fiery,roddick2018orthographic,philion2020lift,reading2021categorical,harley2023simple}. Lift-Splat-Shoot \cite{philion2020lift} demonstrates a method that uses intrinsic and extrinsic camera information to lift and condense predictions into a top-down view map for object detection and semantic segmentation. Such map is later used by a model predictive controller to avoid detected objects. FIERY \cite{hu2021fiery} proposes an improvement for the prediction that not only uses the instantaneous cameras' observations, but a short sequence of images to predict the map and future instances. However, the assignment of predefined semantic classes as navigational costs could be restrictive or inaccurately represent feasibility in a specific environment. Methods supervised by humans often fall short of providing sufficient guidance for effective navigation in intricate and unstructured off-road scenarios. One key reason for this is that humans tend to rely on visual information in segmentation. Still, that is not the complete picture of what the robot proprioception sensors actually experience, and therefore, not a determination of traversability for the robot.

In contrast, \textbf{self-supervised strategies} employ feedback from interactions with the environment to mitigate these limitations \cite{gasparino2022wayfast,bansal2020combining,wellhausen2020safe,wellhausen2019should,gasparino2023unmatched,mishra2021deep,frey2023fast}. Through assimilating data ranging from IMU, traction, RGB-D, and LiDAR regarding a vehicle's interaction with the terrain, these approaches identify traversable areas or classify terrains into categories associated with costs. In \cite{wellhausen2020safe}, anomaly detection utilizing normalizing flow techniques is leveraged to identify out-of-distribution scenarios within the robot experiences. However, this method assumes that the dataset only contains safe interactions and does not use the information from failures. In BADGR \cite{kahn2021badgr}, the authors train an end-to-end policy with automatically created labels using a set of rules to define collisions and bumpiness from embedded sensors. Although this method avoids explicit human labeling, it still requires a person's judgment to define the rules to create the labels. WayFAST \cite{gasparino2022wayfast} addresses these limitations by using a kino-dynamic model allied with a receding horizon estimator (RHE) to label the recorded data. The WayFAST method is able to learn solely from robot experiences and demonstrates navigation in challenging unstructured outdoor environments, such as terrains with snow, sand, and tall grass. WVN \cite{frey2023fast} shows a similar method that adds a mechanism to adapt the network online and train the model while in the exploration phase. Such method can increase the adaptability to new unseen terrains. Yet, to the best of our knowledge, there is not a self-supervised method that can learn to predict traversability by fusing proprioceptive and temporal information to improve navigation robustness.

\section{Contribution: Methodology \& System Design}

\textbf{The main contribution} of this paper is in providing a novel method for traversability-based navigation for outdoor unstructured environments. Our method is completely self-supervised, increasing the easiness of deployment while not needing any manual labels or heuristics to be defined. Different from previous methods \cite{gasparino2022wayfast,kahn2021badgr,frey2023fast}, \method{} predicts the traversability map directly in a BEV map and uses temporal information to keep track of obstacles that are not visible anymore, as illustrated in Fig. \ref{fig:main-image}. We show through experiments that our method is robust and can successfully navigate in challenging scenarios.

\subsection{State Estimation and Automatic Data Labeling} \label{subsection:mhe}

\begin{figure*}[htp]
    \centering
    \includegraphics[trim=0cm 2.5cm 0cm 4cm,clip,width=0.88\linewidth]{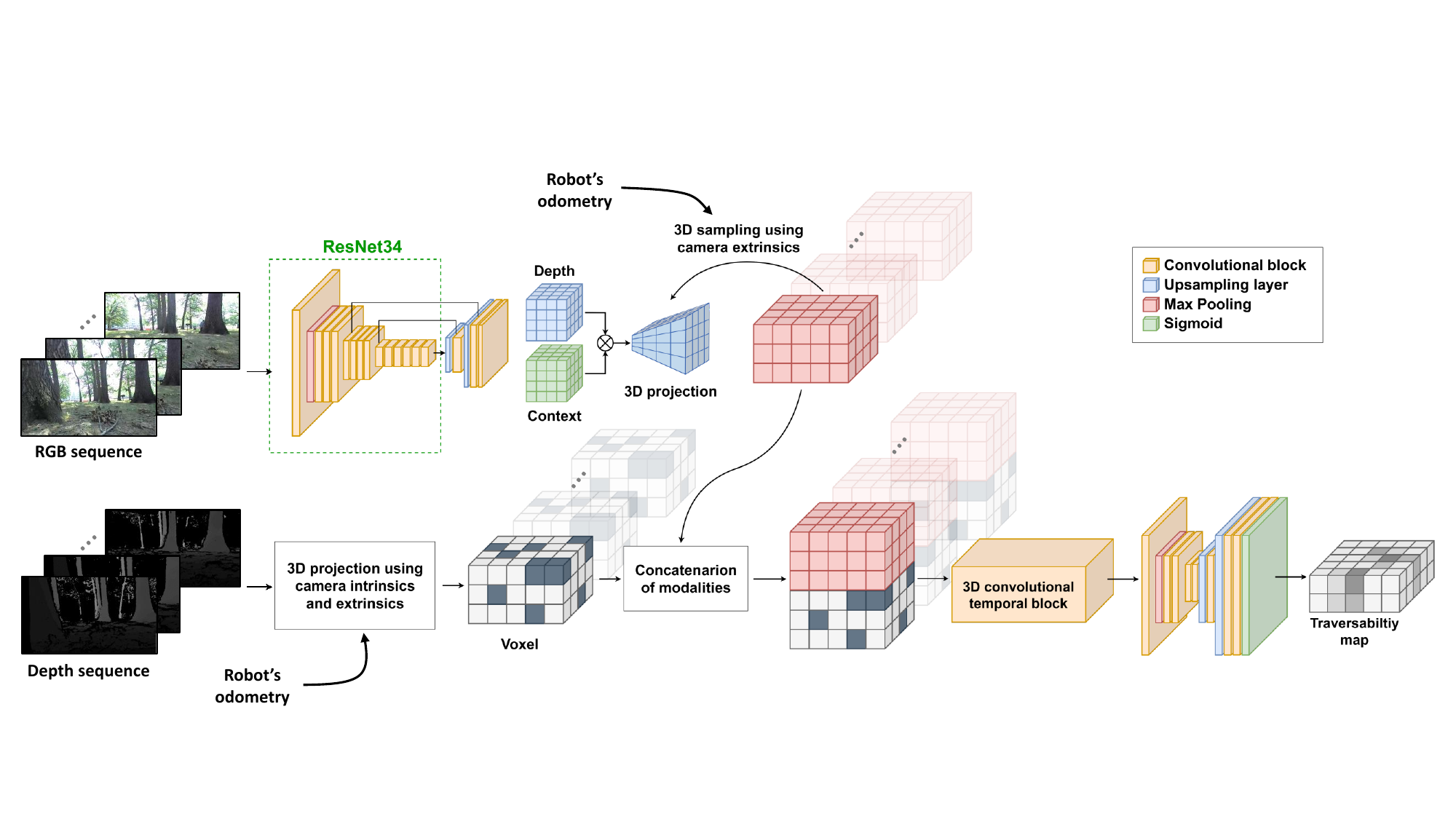}
    \caption{Our network model takes a sequence of RGB and depth images to predict a local traversability map. The depth information is fused explicitly in the form of voxels in the network's latent space. A 3D convolutional block fuses the sequence of concatenations and a final 2D convolutional block outputs the traversability map.} 
    \label{fig:net-architecture}
    \vspace{-0.2in}
\end{figure*}

To train the neural network responsible for predicting the traversability map, we employ an RHE to concurrently estimate the robot's poses in an inertial coordinate frame while generating labels for the network training. Such labels consist of the traversability coefficients $\mu$ and $\nu$, as described in \cite{gasparino2022wayfast}, and denote a coefficient in the control channel of the robot's kino-dynamic model that relates to the terrain traction. They are designed to approach a value of one when the vehicle follows the control actions, and zero when the system fails to comply with the actions.

\vspace{-0.1in}

\begin{equation}
    x_{k+1}
    =
    \begin{bmatrix}
        p_{x_k} \\
        p_{y_k} \\
        \theta_k
    \end{bmatrix}
    +
    \begin{bmatrix}
        \mu \cdot cos(\theta_k) & 0 \\
        \mu \cdot sin(\theta_k) & 0  \\
        0 & \nu
    \end{bmatrix}
    \begin{bmatrix}
        v_k \\
        \omega_k
    \end{bmatrix}
    \cdot \Delta t
    \label{eq:kino-model}
\end{equation}

The estimator uses the discrete system model shown in \eqref{eq:kino-model} and synchronously runs with the GNSS module on the robot at a sampling time $\Delta t$. We solve the following finite horizon optimization formulation to obtain the system states and unknown parameters $\mu$ and $\nu$.

\vspace{-0.1in}

\begin{align} \label{eq:mhe-optimization}
    \begin{gathered}
        \min_{x_{k:k+N}, m} ||x_k-\Tilde{x}_k||^2_{P_x} + 
        ||m-\Tilde{m}||^2_{P_m} + \\
        \sum_{i=k}^{k+N} ||x_i-h(z_i,\Delta\theta)||^2_{P_w}
    \end{gathered}
\end{align}
subject to the constraints
\vspace{-0.1in}
\begin{align*}
    \begin{gathered}
        x_{k+1} = f(x_k,u_k) \\
        \mu, \nu \in [0,1]\textit{, } \Delta\theta \in [-\pi,\pi) \textit{,}
    \end{gathered}
\end{align*}
where $\Tilde{x}_k$ is the initial estimated state vector, $\Delta\theta$ is the offset between true North and compass estimated North, $m=[\mu, \nu, \Delta\theta]^\top$ is the vector of parameters, $\Tilde{m}$ is the estimated vector of parameters from the previous iteration, $N \in \mathbb{N}$ is the length of the estimation horizon, and $P_x$, $P_m$, and $P_w$ are weighing matrices. In Eq. \ref{eq:mhe-optimization}, $h(z_k,\Delta\theta)$ is the measurement equation model where $z_k$ is the sensor measurement that comes from the GNSS (position) and IMU (heading angle). For estimation purposes, we assume the unknown coefficients are constant in a short horizon $N$.

\subsection{Traversability Prediction Module}

In this section, we present the module responsible for generating the traversability map used to guide the robot. The module is composed of a neural network model that takes a sequence of camera observations (color and depth images) and the camera's intrinsic and extrinsic parameters as input to predict a BEV map that contains information about linear and angular traversability. This traversability coefficient represents how easy in terms of control effort it is to traverse that region of the terrain.

As shown in Fig. \ref{fig:net-architecture}, our network uses a ResNet34 backbone \cite{he2016deep}, pre-trained on the ImageNet dataset \cite{deng2009imagenet}. We choose ResNet34 because it is lightweight and can run in real-time on embedded devices. We truncate the ResNet34 network after four convolutional blocks, such that the output is a tensor with 1/16 of the input resolution. After this initial encoder, we add a decoder with two blocks of convolutional layers, such that the latent tensor is fused with features from the encoder and up-sampled to 1/4 of the initial resolution. The output tensor is split into two parts in the channel dimension, creating two latent representations of the same resolution. One of these tensors represents the depth distribution of the camera image, and the other is a latent representation that contains contextual information about the extracted features. Using the latent depth distribution, we propagate each downsampled context pixel information by applying the outer product of such tensors, creating a contextual frustum that represents the extracted features in the 3-dimensional space of that respective camera view, as described in \cite{philion2020lift}. Using the camera extrinsic transformation, we 3D sample the frustum to a latent 3D voxel that represents features extracted around the robot. A second branch applies a simple projection of the depth information, generating an occupancy voxel representation and concatenating it with the features extracted from the color images. Note that both these latent representations describe the same spatial information.

Since the latent voxel describes a single temporal step, the same process is repeated for each observation in the input sequence. Similar to \cite{hu2021fiery}, we use a 3D convolutional block to fuse the sequence of features temporally. The output is then passed through an encoder-decoder block, responsible for fusing the purely geometric information from the depth branch with the contextual voxel from the image branch to output the traversability map. Note that the traversability map is composed of two channels: One for linear traversability, and another for angular traversability.

\subsection{Dataset and Network Training}

To acquire data for training the traversability prediction network, we manually drove the robot in forest-like and semi-urban outdoor environments. We collected camera observations and state estimations using our estimator (described in \ref{subsection:mhe}) for about nine hours, equivalent to a dataset of about 65000 image pairs. This dataset includes the dataset from \cite{gasparino2022wayfast} and new data collected for this article. As shown in \ref{subsection:mhe}, the RHE estimates the robot 2D poses with the traversability values. In addition, we use an inertial sensor to measure the robot's 3D attitude, which is used to generate the camera's extrinsic transformation. Then, from the recorded data, we create a dataset where the data points are tuples of the form $(o_{k-N+1:k},T_{k-N+1:k},K_{k-N+1:k})$ and $(x_{k-M+1:k+M},trav_{k-M+1:k+M})$, where $o_{k-N+1:k}$, $K_{k-N+1:k}$, and $T_{k-N+1:k}$ are synchronized sequences of length $N$ of image observations, camera intrinsic and extrinsic transformations, respectively. Note that the extrinsic transformations are relative to the robot's base pose at time step $k$, and therefore, the robot's pose estimation is incorporated into this transformation. The camera's intrinsics are accurately obtained from its manufacturer. $x_{k-M+1:k+M}$ and $trav_{k-M+1:k+M}$ are the sequences of 2D poses and traversability values of length $2M$ that are used as labels to train the network model.

To train our network, we follow a method similar to \cite{gasparino2022wayfast}, with the difference that instead of projecting the navigated path to the image, we directly sample map values using a differentiable bi-linear interpolation sampling operation. Such operation simplifies the path projection, and therefore, avoids the necessity of defining the thickness of the path as in WayFAST \cite{gasparino2022wayfast}. Similar to \cite{gasparino2022wayfast}, we use a label distribution smoothing technique to balance the training labels according to their frequency of occurrence, based on the approach shown in \cite{yang2021delving}. Since our training data is sparse due to the sparse sampling of traversability values, we use the depth measurements from the depth image to stabilize the training. The idea is to drive the output of the first image encoder to be a distribution of depth values that are near the estimated ones while giving a dense depth prediction since the camera's depth is noisy and not always complete. This method demonstrated better training stability and faster convergence according to our experiments.

To avoid overconfidence in the camera's depth estimation modality, we use a dropout in the depth modality (second branch that fuses visual information with a 3D voxel created from the depth image). During training, the depth modality is zeroed out with a probability of 30\%, to force the network to learn from the colored visual information. Therefore, the final training loss is defined as
\begin{align*}
    \mathcal{L} = &\dfrac{1}{2M}\sum_{i=1}^{2M+1} w_i |trav_i \text{ - } \widehat{trav}_{i} | \text{ +} \\
    &\lambda \cdot \dfrac{1}{W \cdot H} \sum_{u=1}^{W}\sum_{v=1}^{H} L_{CE}(D(u,v), \widehat{D}(u,v)) \text{,}
\end{align*}
where the first term is responsible for training the traversability map output, and the second is for the visual depth prediction. The visual depth prediction is trained as a multi-class classification problem, where each depth discretized bin is treated as a class \cite{reading2021categorical}. For the first term of the loss function, $2M$ is the total number of state/traversability pairs for the sequence, $w_i$ is the weight to balance the traversability dataset distribution, $trav_i$ and $\widehat{trav}_i$ are the traversability output and label for the $i^{th}$ step in the sequence. For the second term, $W$ and $H$ are the width and height of the depth output tensor, $L_{CE}$ is the cross-entropy loss, $D(u,v)$ and $\widehat{D}(u,v)$ are the depth and target depth probabilities for the tensor location $(u,v)$, and $\lambda$ is a weight to balance the training in prioritize the traversability prediction.

\subsection{Traversability-based Model Predictive Controller}

In order to control the robot and guide it to the reference, we use the same kino-dynamic model described in \eqref{eq:kino-model}, but with the difference that $\mu$ and $\nu$ are now functions of the robot's position. These functions come as the output of our neural network model, where the traversability map is used as a lookup table from a bi-linear interpolation. We define these continuous functions as $\mu(p_x,p_y)$ and $\nu(p_x,p_y)$, parameterized as a function of the system states $p_x$ and $p_y$.

\begin{equation*}
    \resizebox{0.95\hsize}{!}{%
        $x_{k+1}
        =
        \begin{bmatrix}
            p_{x_k} \\
            p_{y_k} \\
            \theta_k
        \end{bmatrix}
        +
        \begin{bmatrix}
            \mu(p_{x_k}, p_{y_k}) \cdot cos(\theta_k) & 0 \\
            \mu(p_{x_k}, p_{y_k}) \cdot sin(\theta_k) & 0  \\
            0 & \nu(p_{x_k}, p_{y_k})
        \end{bmatrix}
        \begin{bmatrix}
            v_k \\
            \omega_k
        \end{bmatrix}$
    }
\end{equation*}

We choose an optimization horizon $N \in \mathbb{N}$ and positive definite matrices $Q$, $R$, and $Q_N$. The following finite horizon optimization formulation is solved to obtain a sequence of control actions that minimizes the cost function
\begin{align} \label{eq:optimization}
    \begin{gathered}
        \min_{u_{k:k-N}} \sum_{i=k}^{k+N-1} \left\{ ||x_i-x_i^r||^2_Q + ||u_i-u_i^r||^2_R \right\} \\
         - \sum_{i=k}^{k+N} \left\{ W_\mu \cdot \mu(p_{x_i}, p_{y_i}) + W_\nu \cdot \nu(p_{x_i}, p_{y_i}) \right\} \\
         + ||x_{k+N}-x_{k+N}^r||^2_{Q_N}
    \end{gathered}
\end{align}
where the first line is utilized to drive $x_i$ and $u_i$ to the reference values $x^r_i$ while $u^r_i$; the second line is used to maximize the path traversability; and the last term minimizes the terminal state error.

\begin{figure}[htp]
    \centering
    \includegraphics[trim={3cm 1cm 3cm 1cm},clip,width=0.95\linewidth]{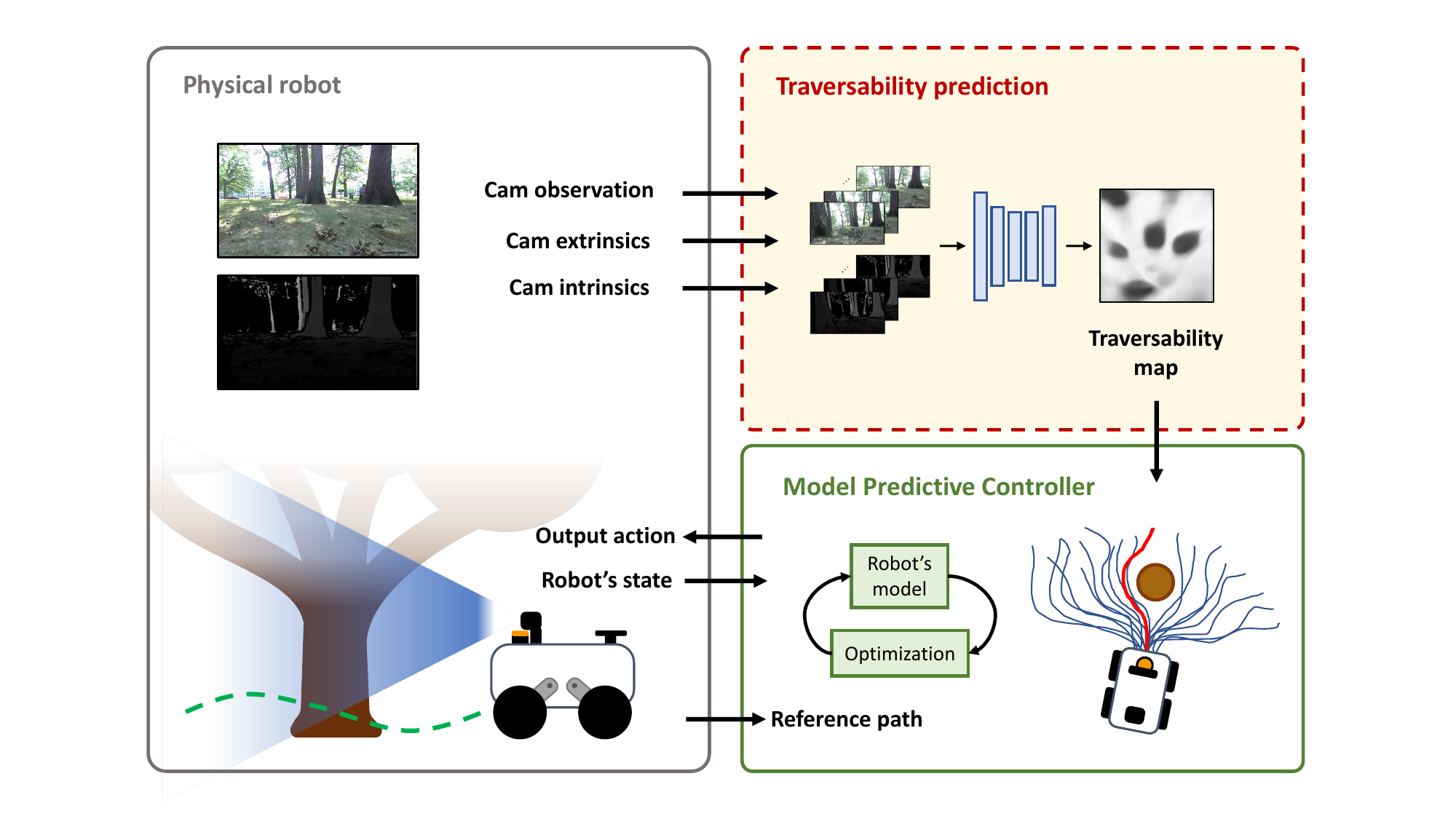}
    \caption{\textbf{Our system architecture}. A sequence of observations is used to predict a local traversability map. The traversability map is directly used by a model predictive control, that samples the map to get the local system parameters.} 
    \label{fig:mpc-diagram}
    \vspace{-0.1in}
\end{figure}

The first control action, solution of \eqref{eq:optimization}, is applied to the system to drive the robot towards the goal. Fig. \ref{fig:mpc-diagram} shows how the predicted traversability map is used in the model predictive control formulation. We utilize a sampling-based controller inspired by the works \cite{williams2017model,williams2017information}, where action trajectories are sampled and evaluated to find the minimum cost path within the camera field of view.

\section{Validation \& Experimental Results}

In this section, we present the experimental results for \method{}. We performed offline experiments on a validation dataset and deployed our method on real robotic platforms to show quantitative and qualitative results of the traversability prediction framework. 

\subsection{Offline validation and ablation}
We performed ablation studies with offline validation to quantify how well the traversability predictions are improved with temporal and geometric voxel fusion. We split the dataset into two parts containing 70\% and 30\% of the data. The first split was used to train the network model, and the second for offline validation. Table \ref{tab:network-baselines} summarizes the results for three variations of our network model. For each baseline, we trained three random weight initializations (except for the ResNet34 pre-trained weights) and averaged the obtained results. The reported results are the average error across the three initializations for the best validation epoch, where the error is calculated as the average of the absolute error between the predicted traversability and experienced traversability (labels created by the RHE) for a given navigated path. According to table \ref{tab:network-baselines}, the depth fusion in the form of voxels and temporal aggregation provides the lowest prediction error, and therefore, we used this method for all the subsequent experiments.


\begin{table}[htp]
    \centering
    \begin{tabular}{cc}
        \hline
        \textbf{Method} & \textbf{Avg. abs. error} \\ \hline
        Vision-only & 0.125 \\
        Voxel fusion & 0.115 \\
        Voxel + temporal fusion & \textbf{0.095} \\ \hline
    \end{tabular}
    \caption{\textbf{Network model validation}. \textit{Vision-only} stands for the network that doesn't use the lower branch that fuses the depth image in the form of voxels. \textit{Voxel fusion} presents the result that fuses the depth in the form of voxels but with a single image as input. And \textit{voxel + temporal fusion} is the whole network as shown in Fig. \ref{fig:net-architecture}.} 
    \label{tab:network-baselines}
    \vspace{-0.2in}
\end{table}

\subsection{Comparison with baselines in real outdoor environment} \label{subsection:line-experiment}

In this experiment, we compared the navigation performance of our method \method{} against two strong baselines: WayFAST \cite{gasparino2022wayfast}, and LiDAR-based navigation in real outdoor environments. We used a TerraSentia robot \cite{gasparino2022wayfast,gasparino2023cropnav}, manufactured by EarthSense Inc., equipped with a StereoLabs ZED2 camera that provides the RGB and depth images used in this work. The goal is to navigate in a challenging environment with tall grass, corridors, and sharp turns while passing through three intermediate waypoints to reach the final goal waypoint. The environment is considered challenging because the robot has to cross an area with dense tall grass of the same height as the robot, circumvent a building, perform a sharp turn while entering the building, and reach the goal location. The waypoints are generated by creating points in the inertial coordinate frame, and in all the runs, the robot starts from the same point. For localization, the robot uses the RHE described in \ref{subsection:mhe}. 

\begin{figure}[htp]
    \begin{subfigure}[h]{0.32\linewidth}
        \includegraphics[trim=16cm 2.6cm 15cm 1cm,clip,width=\textwidth]{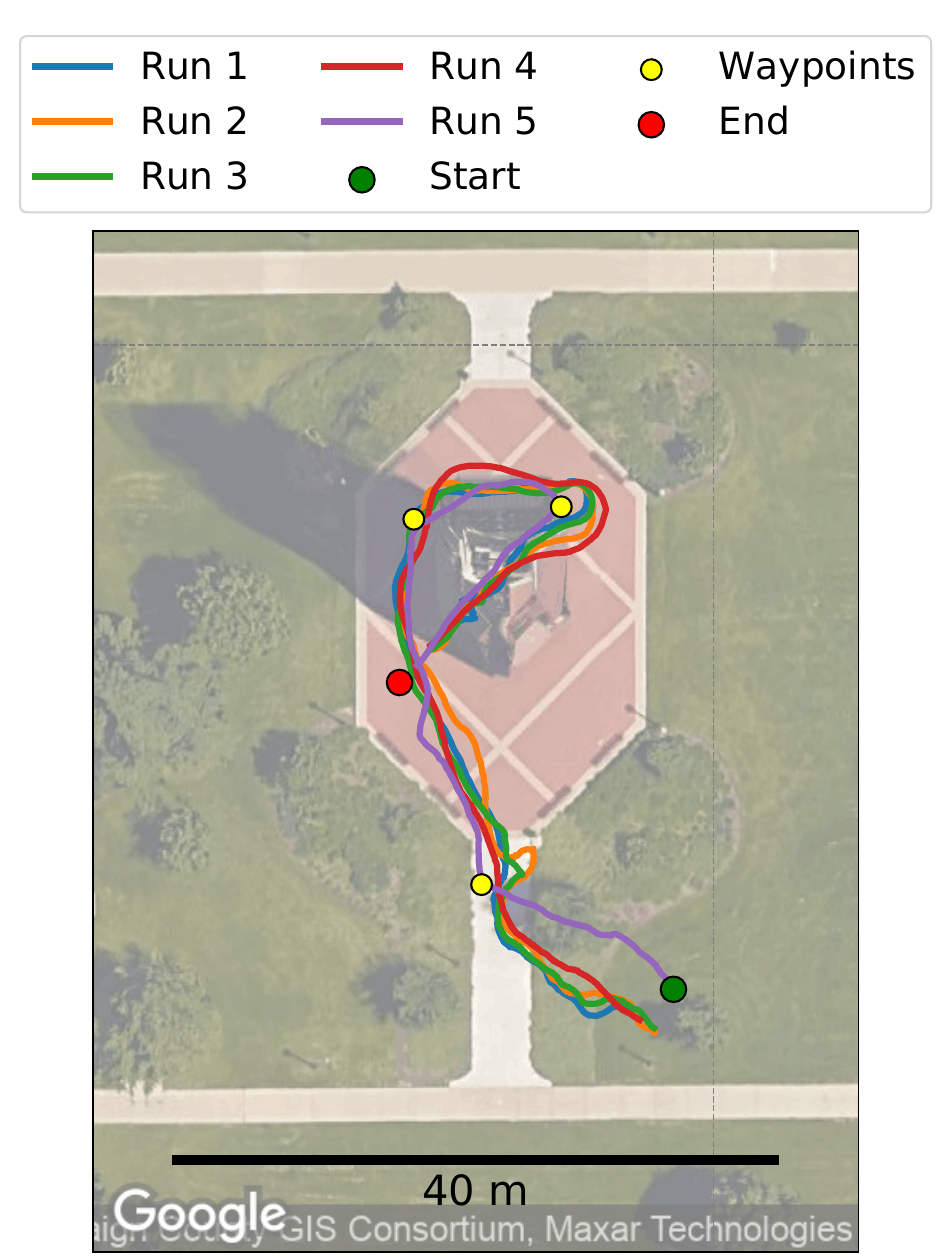}
        \caption*{\footnotesize \textbf{WayFASTER}}
    \end{subfigure}
    \hfill
    \begin{subfigure}[h]{0.32\linewidth}
        \vspace{0.65cm}
        \includegraphics[trim=16cm 2.6cm 15cm 4.65cm,clip,width=\textwidth]{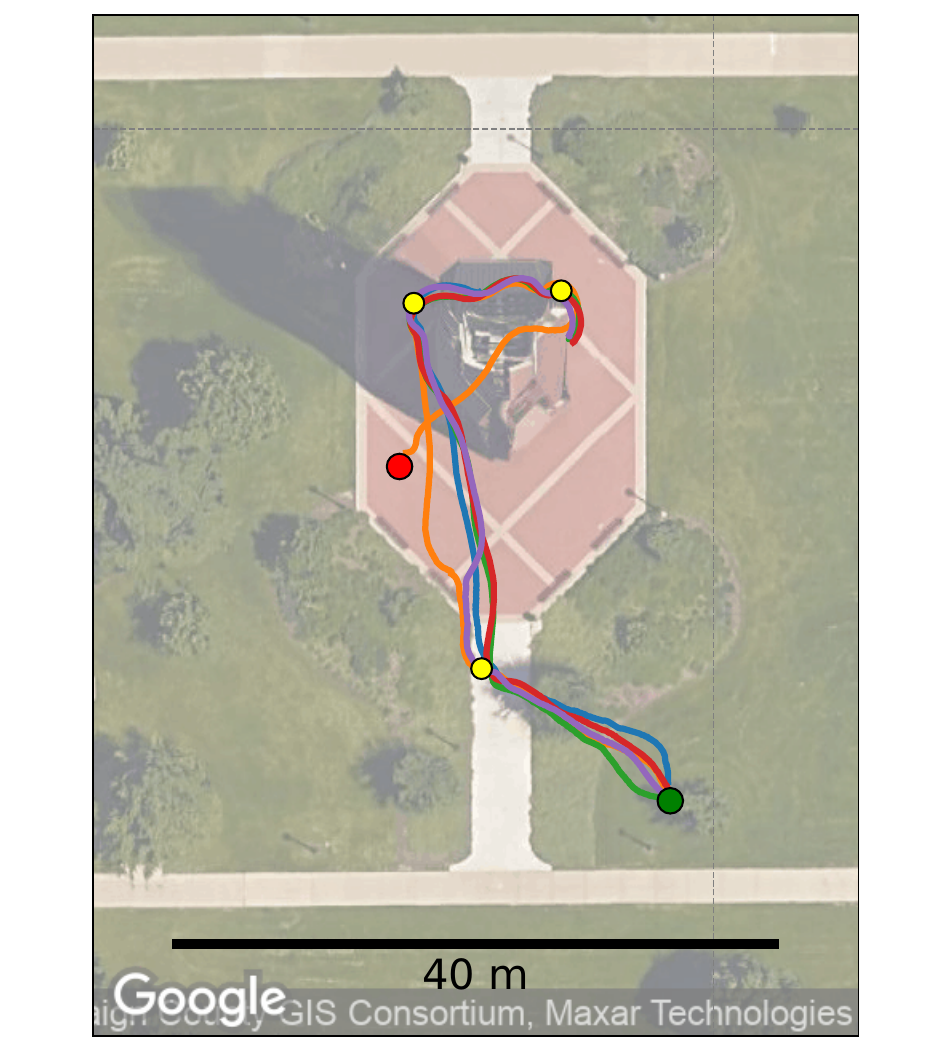}
        \caption*{\footnotesize \textbf{WayFAST}}
    \end{subfigure}
    \hfill
    \begin{subfigure}[h]{0.32\linewidth}
        \vspace{0.65cm}
        \includegraphics[trim=16cm 2.6cm 15cm 4.65cm,clip,width=\textwidth]{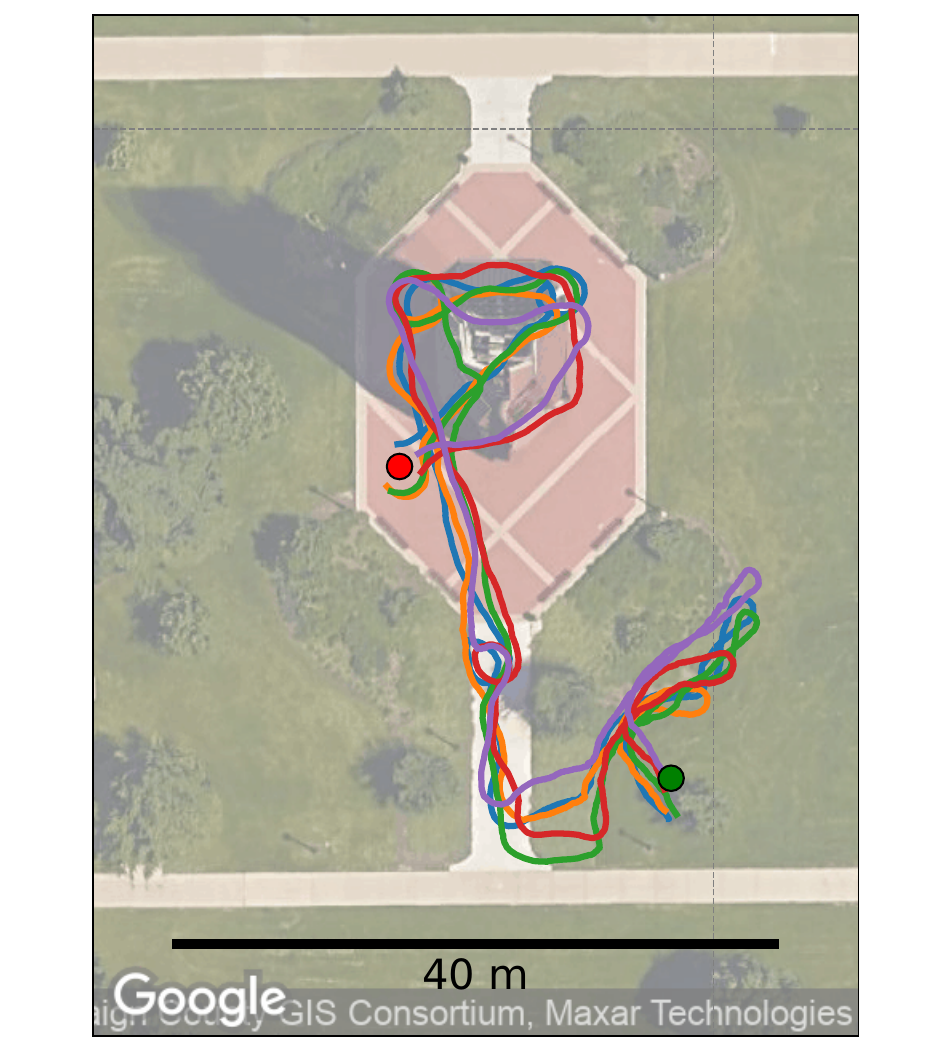}
        \caption*{\footnotesize \textbf{LiDAR-based}}
    \end{subfigure}%
    \caption{Experiment in a challenging environment with tall grass and sharp turns. The first plot shows our method, \method{}. The plot on the middle a navigation using WayFAST \cite{gasparino2022wayfast}, and on the right, the LiDAR-based navigation.} 
    \label{fig:experiments-belltower}
    \vspace{-0.1in}
\end{figure}

For this experiment, we used WayFASTER, WayFAST \cite{gasparino2022wayfast}, and a LiDAR-based navigation as the baselines. The LiDAR-based navigation employs the same framework as WayFASTER, but with a key difference: the traversability map is directly generated from a Hokuyo UST-10LX LiDAR measurements. In areas where a LiDAR beam intersects an obstacle, the corresponding grid space is assigned a traversability value of zero, indicating non-traversable terrain, whereas all other areas are designated as traversable with a value of one. All the baselines used the same set of hyper-parameters and the results are summarized in Table \ref{tab:results-belltower}.


\begin{table}[htp]
    \centering
    \begin{tabular}{ccc}
        \hline
        \textbf{Method} & \textbf{Success} & \textbf{Avg. time (s)} \\ \hline
        LiDAR-based & 5/5 & 201 \\
        WayFAST \cite{gasparino2022wayfast} & 1/5 & - \\
        \textbf{\method{}} & \textbf{5/5} & \textbf{118} \\ \hline
    \end{tabular}
    \caption{Success rate and average time for the navigation in a challenging environment with tall grass and sharp turns.} 
    \label{tab:results-belltower}
    \vspace{-0.1in}
\end{table}

As demonstrated in Fig. \ref{fig:experiments-belltower}, \method{} was the most successful method among the three. The WayFAST \cite{gasparino2022wayfast} approach could not reach the final goal in most of the experiments, often failing when performing the final sharp turn. We hypothesize that, since the method could only use a narrow field-of-view, it was not able to respond in time when doing sharps turns since obstacles would suddenly disappear and reappear. The LiDAR-based navigation and our method \method{} were able to successfully complete the whole path. This is because both these methods have a wider field of view. The LiDAR sensor covers 270\textsuperscript{o} of view, and \method{} has temporal fusion to enhance the visibility. However, as we can see in the third image in Fig. \ref{fig:experiments-belltower}, the LiDAR-based navigation could not navigate through the tall grass, which is expected for sensors that rely solely on geometric features. Because of that, the robot had to deviate from the tall grass area, taking much longer to reach the final destination.

\subsection{Qualitative evaluation of traversability prediction maps}

\begin{figure}[htp]
    \centering
    \includegraphics[width=0.85\linewidth]{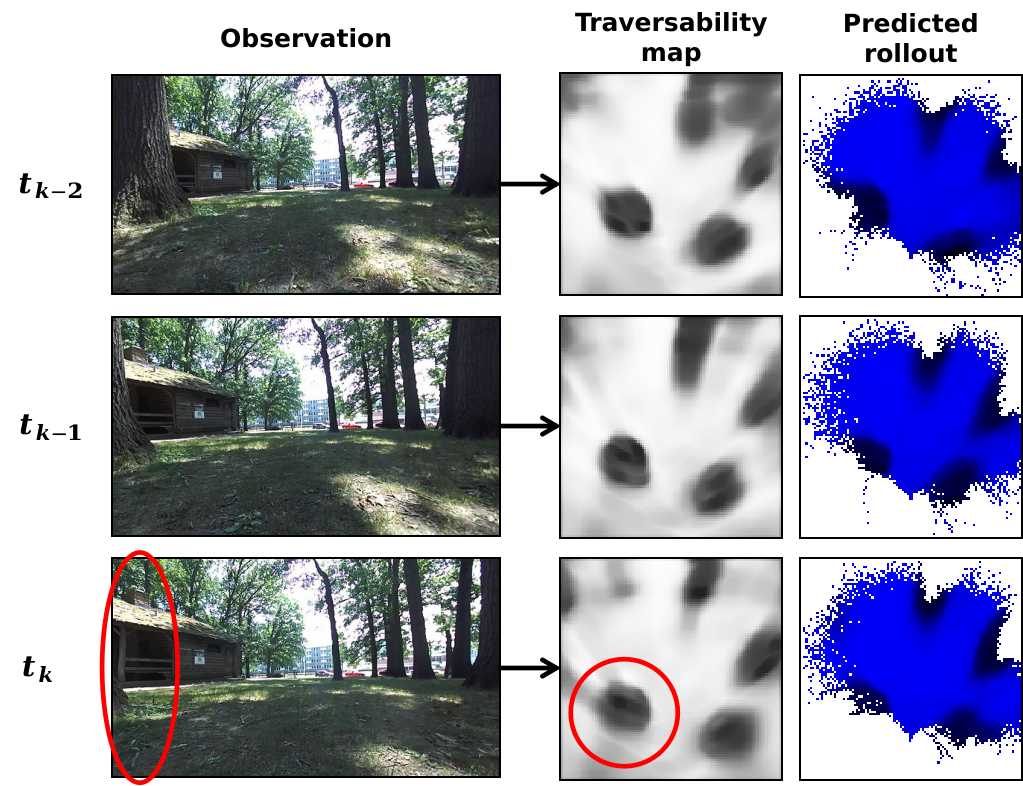}
    \caption{\textbf{Example of temporal prediction}. For each observation on the left, we show the traversability map prediction and the model predictive controller rollouts. Note that the tree is not visible at time $t_k$. Due to the temporal fusion, the predicted map still keeps track of past occurrences.}
    \label{fig:temporal-prediction}
    \vspace{-0.1in}
\end{figure}

Fig. \ref{fig:temporal-prediction} shows an example of visualization of traversability predictions from our method during the experiment. The middle column shows the neural network outputs, where the traversability map keeps track of untraversable areas using temporal fusion even when they are not visible in the current observation anymore. Also, due to the use of traversability as a kino-dynamic model parameter, the controller rollouts bend when they reach untraversable areas. This happens because the untraversable areas cause the model to become uncontrollable, preventing the robot from moving further.  As demonstrated in the \textit{predicted rollouts} column, the model predictive control rollouts can still predict the unseen trees, even after they disappear. Such predictions ensure that the robot stays aware and avoids such obstacles even when they are not immediately seen.

\vspace{-0.1in}

\begin{figure}[htp]
    \includegraphics[width=\linewidth]{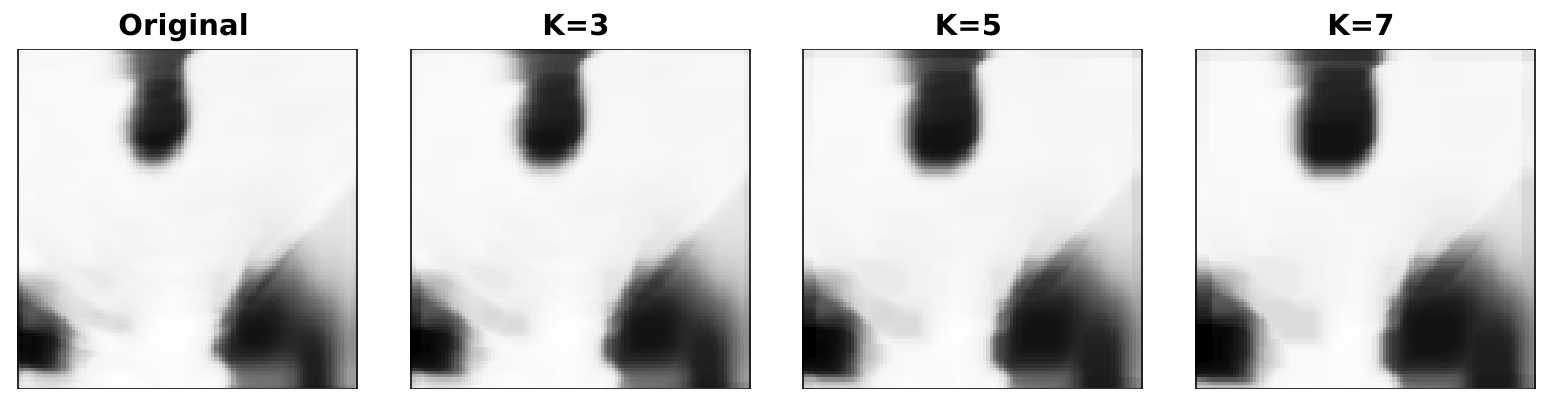}
    \caption{Terrain clearance by kernel size for extracting minimum traversability values. The image shows the original map and modifications with kernels of size 3, 5, and 7.}
    \label{fig:clearance}
    \vspace{-0.1in}
\end{figure}

Another advantage of our BEV traversability map is that it is simple to increase obstacle clearance. As demonstrated in Fig. \ref{fig:clearance}, by applying a convolution kernel for extracting minimum traversability values, we can get different levels of obstacle clearance. Such operation can be easily and efficiently implemented using a $maxpool$ layer function.

\subsection{Long path navigation}

\begin{figure}[htp]
    \begin{subfigure}[h]{0.49\linewidth}
        \centering
        \frame{\includegraphics[trim=1cm 2.5cm 1cm 0cm,clip,width=0.85\linewidth]{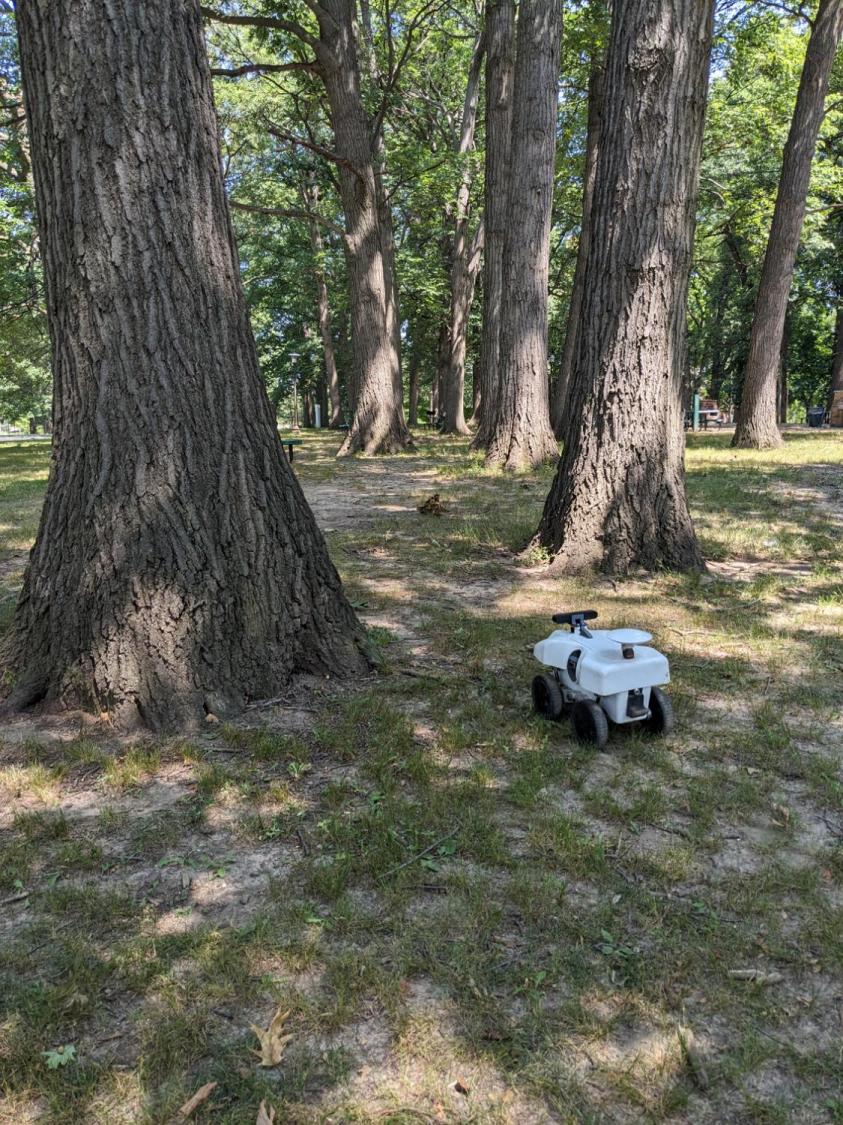}}
    \end{subfigure}
    \hfill
    \begin{subfigure}[h]{0.49\linewidth}
        \centering
        \includegraphics[trim=16cm 2cm 15cm 1cm,clip,width=0.9\linewidth]{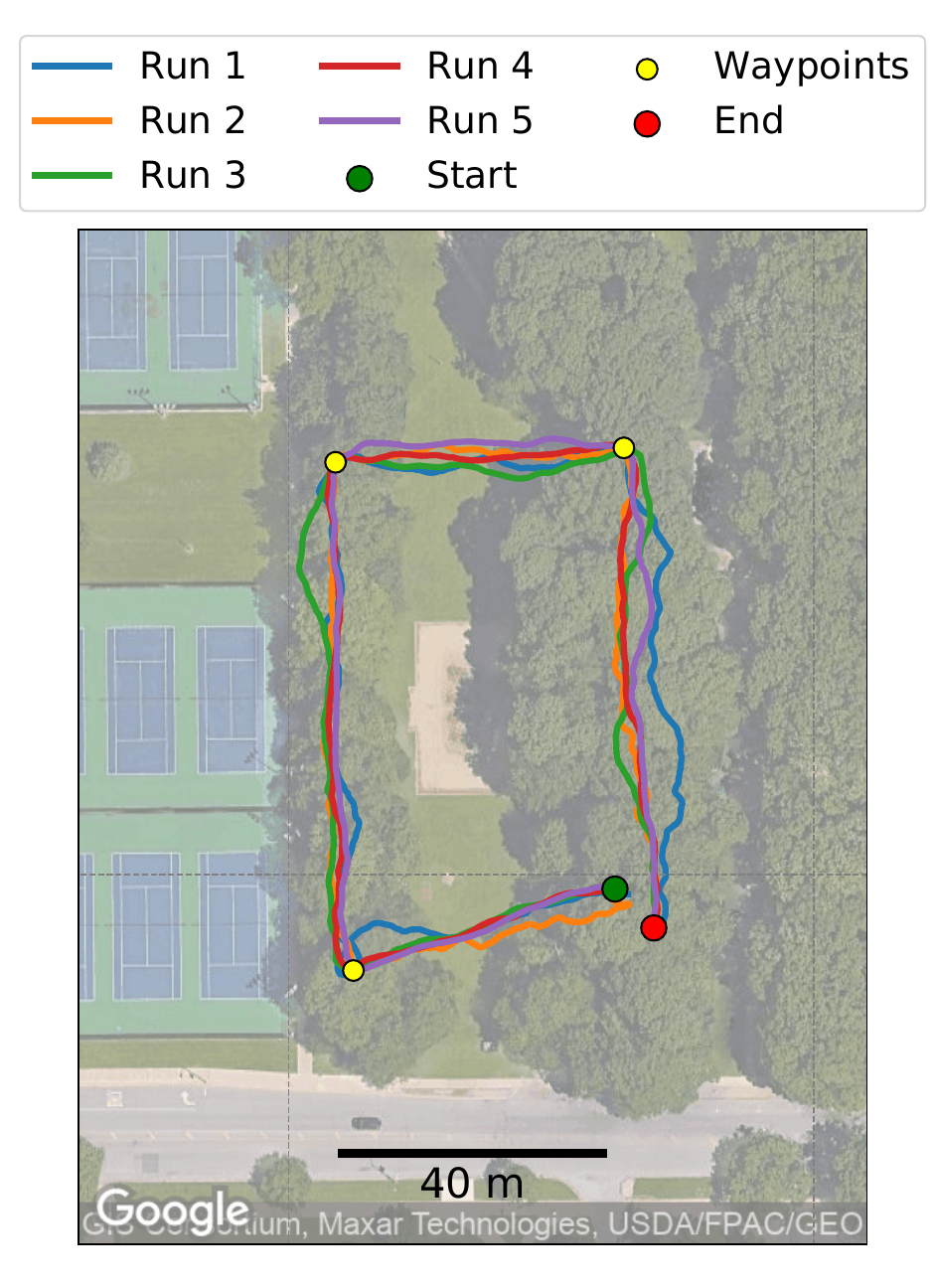}
    \end{subfigure}
    \caption{\textbf{Long path navigation} using our method. We specified four waypoints forming a square and the robot needs to follow these goals while finding a safe path to navigate.} 
    \label{fig:long-path}
    \vspace{-0.1in}
\end{figure}

In this experiment, we show robust navigation of the TerraSentia robot with \method{} on a long path in a forest-like environment. The environment consisted of many trees and natural challenges. The total path was about 180~m long and was constructed using four waypoints. We tested our method five times in this path, and Fig. \ref{fig:long-path} demonstrates this experiment where the robot was successful in all of the runs and safely reached the goal location.

\vspace{-0.1in}

\subsection{Cross-platform validation}

We also tested our method on a different robotic platform. In this experiment, the objective was to show that our modular approach can be easily deployed on different robotic platforms with minor modifications in the hyper-parameters. To do so, we tested \method{} on a Unitree Go1 quadruped robot equipped with a Nvidia Jetson Xavier AGX, a GNSS module, and a StereoLabs ZED2i camera. Since the platform is a quadruped robot, it does not suffer from the same friction when turning as a skid-steer wheeled platform. Therefore, we multiplied the output traversability map channel responsible for the angular traversability prediction by 2.5. We obtained this value by using the RHE module on the TerraSentia robot and taking $1/\nu$ when turning on a flat surface where the traversability is expected to be maximal. This step was enough to match the kino-dynamic model between the quadruped and the wheeled robot.

\vspace{-0.1in}

\begin{figure}[htp]
    \centering
    \begin{subfigure}[h]{0.3\linewidth}
        \centering
        \frame{\includegraphics[width=\textwidth]{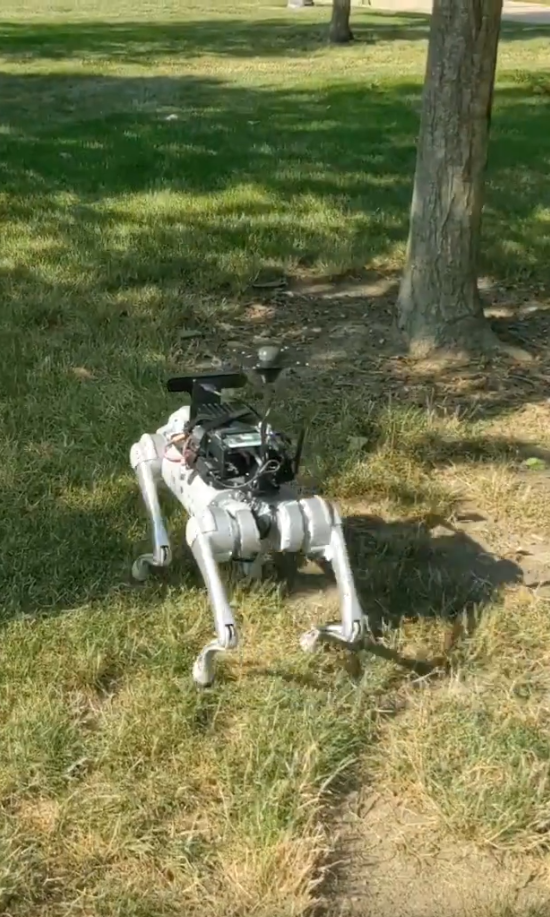}}
    \end{subfigure}
    \hfill
    \begin{subfigure}[h]{0.685\linewidth}
        \centering
        \includegraphics[trim=10cm 2cm 10cm 2cm,clip,width=\linewidth]{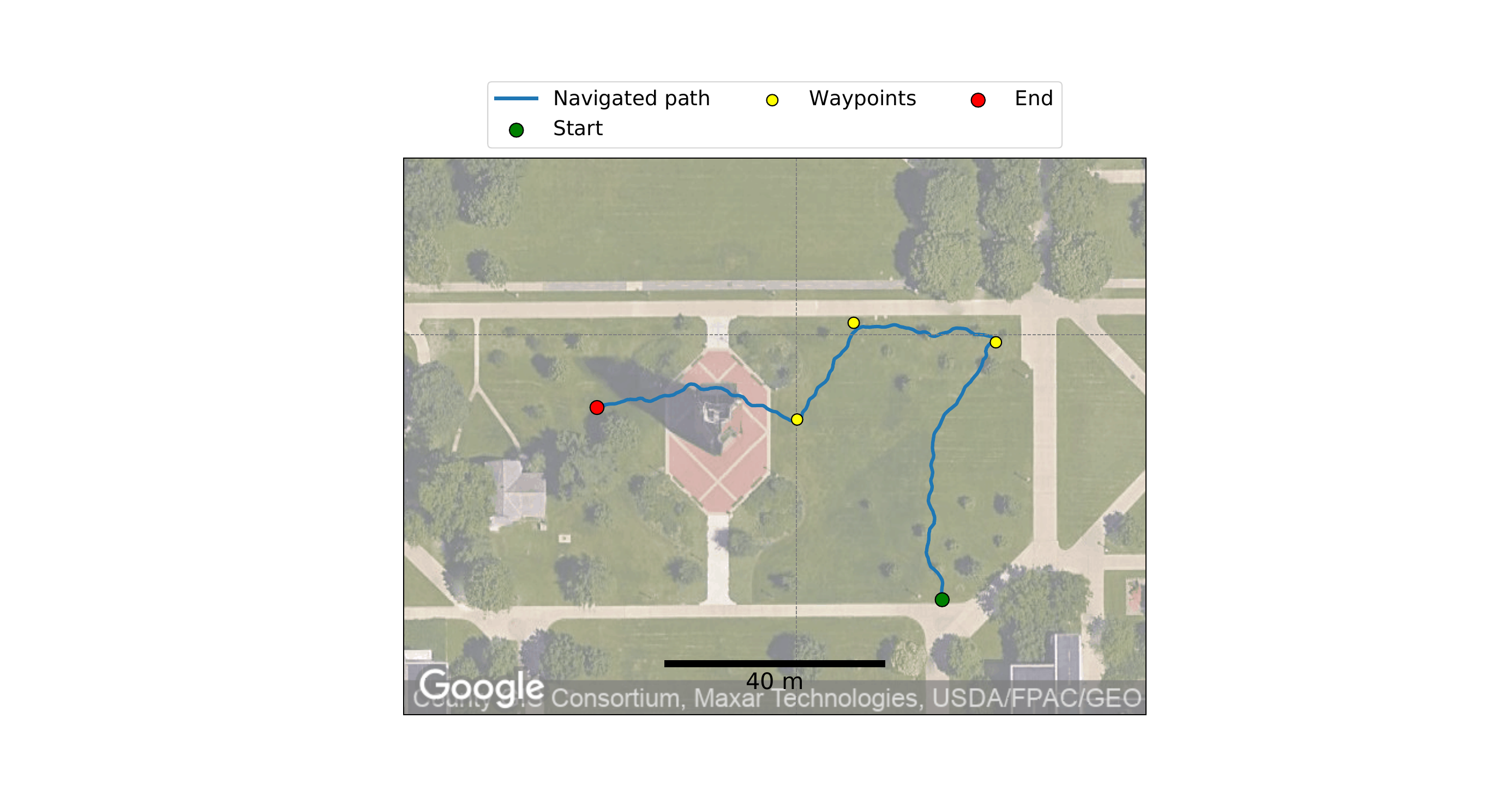}
    \end{subfigure}%
    \caption{Navigation test with \method{} running on a quadruped robot.}
    \label{fig:ours-quadruped}
    \vspace{-0.1in}
\end{figure}

We tested our method on this platform for three runs and our navigation system was able to successfully drive the robot and reach the goal location in all of the tries. Fig. \ref{fig:ours-quadruped} shows one of the experiments where the quadruped was used to navigate a long path using \method{}. As we can see in the image, the robot was able to successfully navigate by avoiding obstacles in between the points, such as trees and buildings, and safely reach all the defined goals.

\section{Conclusion}

We presented \method{}, a novel method for self-supervised traversability estimation that uses sequential information to predict a map that improves the traversability map visibility. For such, we use a neural network model that takes a sequence of RGB and depth images as input, and uses the camera's intrinsic and extrinsic parameters to project the information to a 3D space and predict a 2D traversability map. As we showed through experiments, our system is able to safely guide a robotic platform in a variety of environments and beat the baselines when a wider field of view is required. In addition, we demonstrated that this solution can be easily applied to different platforms with minor modifications, such as a quadruped robot. Furthermore, since \method{} has the receding horizon estimator running at real-time, we believe that, for future work, it is possible to use this information on the fly to adapt the network model, which could further improve the deployment of \method{} to newly unseen environments.

\addtolength{\textheight}{-6cm}   

\bibliographystyle{IEEEtran}
\bibliography{references}

\end{document}